\documentclass[10pt,twocolumn,letterpaper]{article}
\usepackage[pagenumbers]{cvpr}
\usepackage{tikz}
\usetikzlibrary{arrows.meta,positioning,calc}
\usepackage{pgfplots}
\usepgfplotslibrary{groupplots}
\pgfplotsset{compat=1.18}
\usepackage{booktabs}
\usepackage{multirow}

\definecolor{cvprblue}{rgb}{0.21,0.49,0.74}
\usepackage[breaklinks,colorlinks,allcolors=cvprblue]{hyperref}

\title{Hypothesis--Evidence Interaction for Localizing Open-Ended Image Forgeries}
\author{%
    Xiaoxia Chen
    \quad
    Leyu Zhou
    \quad
    Jingyi Hou
    \quad
    Zhijie Liu
    \\
    University of Science and Technology Beijing
}
\date{}
\hypersetup{
    pdftitle={Hypothesis--Evidence Interaction for Localizing Open-Ended Image Forgeries},
    pdfauthor={Xiaoxia Chen, Leyu Zhou, Jingyi Hou, Zhijie Liu},
    pdfdisplaydoctitle=true
}

\begin{document}
\maketitle
\begin{abstract}
As generative models and editing tools continue to evolve, image forgeries become more diverse in their manipulation patterns and forensic traces.
Although these forgeries can still leave useful forensic evidence, the traces are often weak and unevenly reliable across regions, making accurate localization difficult.
Revising a prediction may recover missed manipulations but can also introduce new errors, making it necessary to assess whether the available evidence supports each revision.
We propose a hypothesis--evidence interaction framework that treats a localization map as a hypothesis to be progressively evaluated and revised using forensic evidence.
A contextual candidate generator proposes revised localization maps, while a shared Hypothesis--Evidence Assessment module learns to compare the current and candidate hypotheses using the same forensic evidence.
Their relative support guides selective updates, so each revision depends on how well the proposed change agrees with the evidence in the input image.
The updated hypothesis then informs subsequent generation and assessment, allowing localization decisions to be reconsidered over multiple steps under diverse manipulation conditions.
Experiments on conventional and AI-generated manipulation benchmarks show improved localization, with larger gains on AI-generated forgeries that reflect the evolving diversity and open-ended nature of image manipulation, demonstrating the effectiveness of hypothesis--evidence interaction for open-ended forgery localization.
\end{abstract}

\section{Introduction}

Advances in generative AI have made image manipulation more realistic, raising growing concerns about image authenticity.
As editing techniques continue to evolve, image forgeries become harder to describe through a fixed set of manipulation patterns.
This makes image forgery localization (IFL) increasingly open-ended: evolving manipulation techniques may produce patterns beyond those represented in the training data.
Manipulated regions can blend naturally into the surrounding image, while the forensic traces they leave may be weak and unevenly reliable.
Reliable localization therefore requires careful interpretation of the available evidence to identify edited regions.

Recent progress in IFL has largely focused on learning more informative representations of forensic evidence~\cite{DBLP:conf/iccv/ChenDJC021,DBLP:journals/pami/DongCHCL23,DBLP:conf/cvpr/LiMGXL024}.
Noise-sensitive and compression-aware methods use changes in noise statistics or compression artifacts as evidence of manipulation~\cite{DBLP:conf/cvpr/0001AN19,DBLP:conf/wacv/KwonYNL21,DBLP:journals/ijcv/KwonNYLK22}.
Boundary-aware methods model manipulation boundaries~\cite{DBLP:conf/cvpr/LiZ00FZ23,DBLP:journals/tcsv/ChenCWLCLZW25}, while structural approaches capture inconsistencies between image regions~\cite{DBLP:conf/cvpr/WangWCHSLJ22,DBLP:journals/pami/HanWBWHX24}.
Frequency analysis~\cite{DBLP:conf/ijcai/LiZLLF0LZ25} and mesoscopic modeling~\cite{zhu2025mesoscopic} further broaden the evidence available for localization.
These advances strengthen forensic representations, yet weak manipulation traces can still produce ambiguous feature responses.
How these responses are interpreted at the prediction stage also affects localization reliability.

In many existing pipelines, forensic features are directly mapped to pixelwise predictions through a one-shot prediction head.
Building on these predictions, progressive updating can use the current localization map as context for subsequent refinement, allowing ambiguous responses to be reassessed over multiple steps.
Uncertainty-aware localization methods~\cite{ji2023uncertainty} model prediction reliability, while sequential decision methods~\cite{peng2024code} explicitly update localization maps over multiple steps.
These studies motivate treating the localization map as an intermediate hypothesis that can be progressively revised using the available evidence.
However, a candidate revision may recover a missed manipulated region while also introducing false positives or suppressing a correct response.
Prediction confidence alone cannot reliably indicate whether the evidence supports such a change.
This motivates evaluating a candidate relative to the current prediction before adopting its changes.

\begin{figure}[t]
\centering
\includegraphics[width=\columnwidth]{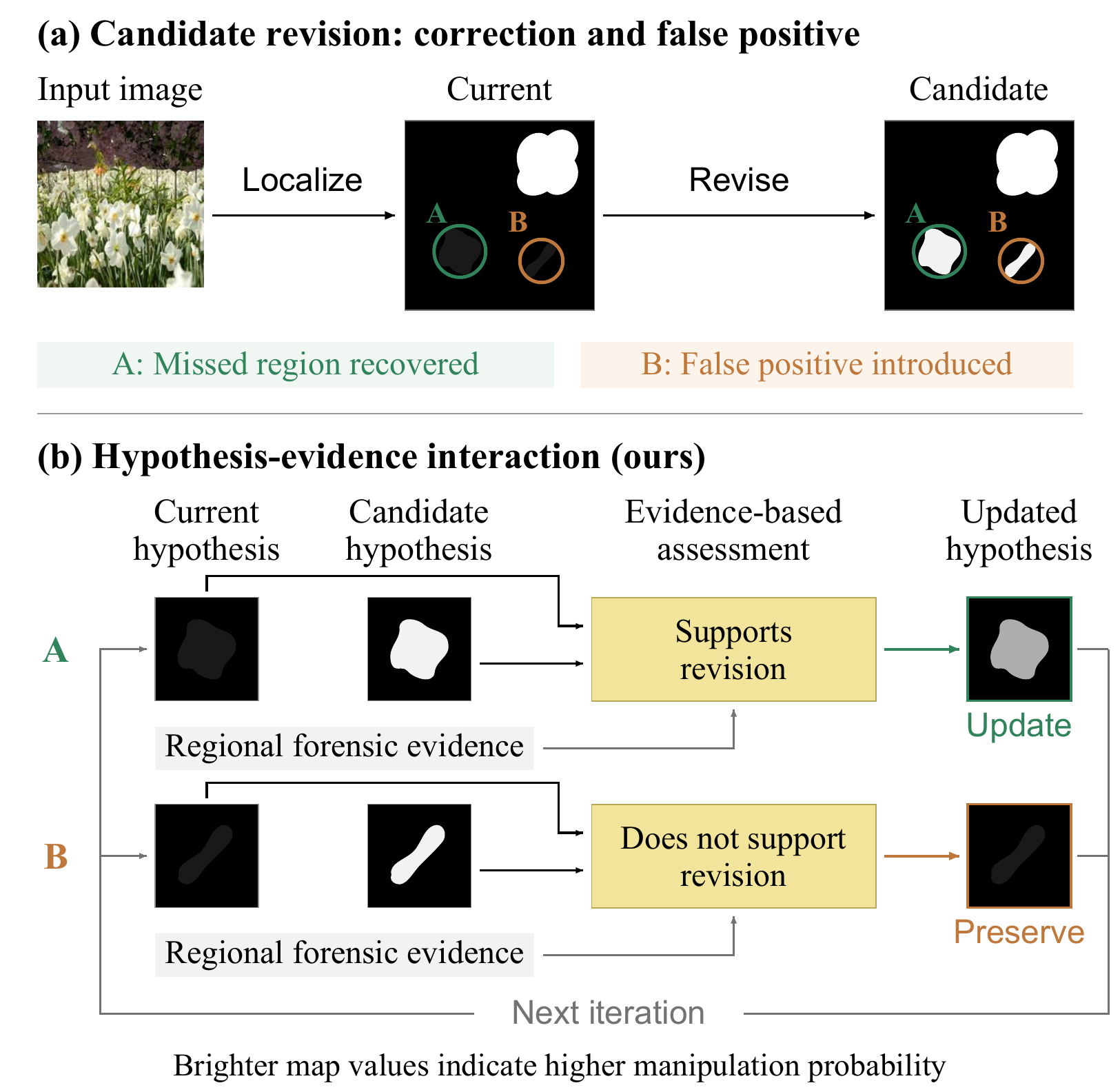}
\caption{
Schematic illustration of hypothesis--evidence interaction for forgery localization.
(a) A candidate revision recovers a missed manipulated region (A) but introduces a false positive (B).
(b) The current and candidate localization maps are treated as alternative hypotheses and assessed using the same regional forensic evidence.
In this example, the evidence supports the revision in A but not in B, guiding a partial update in A while preserving the current prediction in B.
The updated hypothesis becomes the current hypothesis for the next iteration.
}
\label{fig:motivation}
\end{figure}

The current and candidate localization maps represent alternative hypotheses about the manipulated regions.
Comparing their compatibility with the same forensic evidence provides a basis for assessing proposed changes, including corrections to confident but incorrect predictions.
We therefore propose a hypothesis--evidence interaction framework for localizing open-ended image forgeries, as illustrated in Fig.~\ref{fig:motivation}.
The framework uses the evidence in each image to assess revisions under varying manipulation conditions, while preserving the current prediction where a candidate lacks positive relative support.
As the hypothesis changes, the same evidence is considered in the context of the revised localization.

Specifically, a Mamba-based candidate generator~\cite{gu2024mamba} uses mesoscopic features and the current localization map to propose contextual revisions.
A shared Hypothesis--Evidence Assessment (HEA) module jointly processes the evidence with each hypothesis and compares their relative support.
The generator receives direct localization supervision, while HEA learns from whether a candidate improves the current prediction.
The comparison determines how much of each candidate revision is adopted, and the updated hypothesis provides the context for the next iteration.
Source code is included in the supplementary material and will be made publicly available.

Our main contributions are summarized as follows:
\begin{itemize}
    \item We address the difficulty of progressive forgery localization when candidate revisions may correct existing errors while introducing new ones. Our hypothesis--evidence interaction framework evaluates proposed revisions using the available forensic evidence to guide selective updates.
    
    \item We develop a shared Hypothesis--Evidence Assessment module coupled with a contextual candidate generator. Their interaction guides selective updates according to the relative evidence support for each revision.
    
    \item We evaluate the framework on conventional and AI-generated manipulation benchmarks to study localization performance across diverse manipulation conditions.
\end{itemize}

\section{Related Work}

\subsection{Forensic Representation for IFL}
\label{sec:rw_representation}

Existing IFL methods mainly differ in how they represent forgery traces. ManTra-Net~\cite{DBLP:conf/cvpr/0001AN19} learns manipulation-sensitive local anomalies, while MVSS-Net~\cite{DBLP:conf/iccv/ChenDJC021,DBLP:journals/pami/DongCHCL23} combines noise and boundary information with multi-scale supervision. ObjectFormer~\cite{DBLP:conf/cvpr/WangWCHSLJ22} models object-level inconsistency. UnionFormer~\cite{DBLP:conf/cvpr/LiMGXL024}, Mesorch~\cite{zhu2025mesoscopic}, and Co-Transformers~\cite{DBLP:conf/aaai/ZhangFWKYN26} further combine RGB, noise, object-aware, and mesoscopic features. F2D-Net~\cite{DBLP:conf/ijcai/LiZLLF0LZ25}, Pixel-Inconsistency Modeling~\cite{DBLP:journals/pami/KongLWLRK25}, DiRLoc~\cite{DBLP:journals/tdsc/ShengQLCH25}, and HDF-Net~\cite{DBLP:journals/pami/HanWBWHX24} reduce the influence of image semantics through frequency decomposition, pixel inconsistency, feature disentanglement, or regional homogeneity. GSRNet~\cite{DBLP:conf/aaai/ZhouCHNSLD20} uses multi-scale refinement, while the Edge-aware Regional Message Passing Controller~\cite{DBLP:conf/cvpr/LiZ00FZ23}, EAN~\cite{DBLP:journals/tcsv/ChenCWLCLZW25}, and M2SFormer~\cite{nam2025m2sformer} emphasize boundary information and spatial details.
Recent work has extended IFL to manipulations produced by generative models. MIML~\cite{DBLP:conf/cvpr/QuZLXPGJ24} and GIM~\cite{DBLP:conf/aaai/ChenHZLZYLC0YLH25} provide large-scale data and evaluation settings for modern and generative manipulations. Detective SAM~\cite{lek2026detective} combines perturbation-derived forensic clues with a segmentation foundation model. FakeShield~\cite{DBLP:conf/iclr/XuZLTH025} and ManipShield~\cite{DBLP:journals/corr/abs-2511-14259} further introduce multimodal reasoning for joint detection, localization, and explanation.
Building on these advances, our method focuses on how the available forensic features can be used to assess proposed localization revisions. In particular, we compare the evidence support for the current and candidate localization maps before updating the prediction.

\subsection{Iterative Forgery Localization and Prediction Reliability}
\label{sec:rw_progressive}

Recurrent Models of Visual Attention~\cite{mnih2014recurrent}, Glance and Focus Networks~\cite{huang2023glance}, and PolyFormer~\cite{liu2023polyformer} study iterative prediction in visual attention, dynamic recognition, and segmentation, respectively. In IFL, CoDE~\cite{peng2024code} formulates localization as a Markov decision process and updates pixel-wise forgery probabilities through reinforcement learning. InpDiffusion~\cite{wang2025inpdiffusion} and ForgDiffuser~\cite{wang2025forgdiffuser} generate localization masks through iterative denoising. For contextual feature extraction, LoMa~\cite{lou2024loma} and ForMa~\cite{guo2025forma} use state-space models to capture long-range dependencies in forensic features.
Prediction reliability has also been considered in forgery localization. Uncertainty-guided Learning~\cite{ji2023uncertainty} estimates data and model uncertainty to improve manipulation detection, while UGEE-Net~\cite{hao2024ugee} combines uncertainty estimation with edge enhancement and cross-level feature propagation. These methods use uncertainty to identify unreliable or difficult predictions.
Iterative methods provide mechanisms for generating or refining a localization map, whereas uncertainty-aware methods estimate the reliability of a prediction. Our method focuses on the relative quality of a proposed revision: the current and candidate maps are evaluated against the same forensic evidence, and their relative support determines whether and to what extent the revision is adopted. This assessment learns whether a candidate improves the current prediction, rather than estimating uncertainty in either prediction alone.

\section{Method}
\label{sec:method}

\subsection{Overview}
\label{sec:overview}

We consider open-ended image forgery localization, where manipulation patterns continue to evolve with advances in generative AI and editing tools.
Under these conditions, we treat a localization map as a hypothesis about which regions have been manipulated and progressively revise it using the available forensic evidence.
The central difficulty is deciding whether a proposed revision improves the current result: contextual refinement may recover missed manipulations, but it can also introduce false positives or remove correct predictions.
Since prediction confidence alone does not reliably indicate whether the evidence supports a revision, we explicitly compare the current and proposed hypotheses against the same image evidence.

\begin{figure*}[t]
\centering
\includegraphics[width=\textwidth]{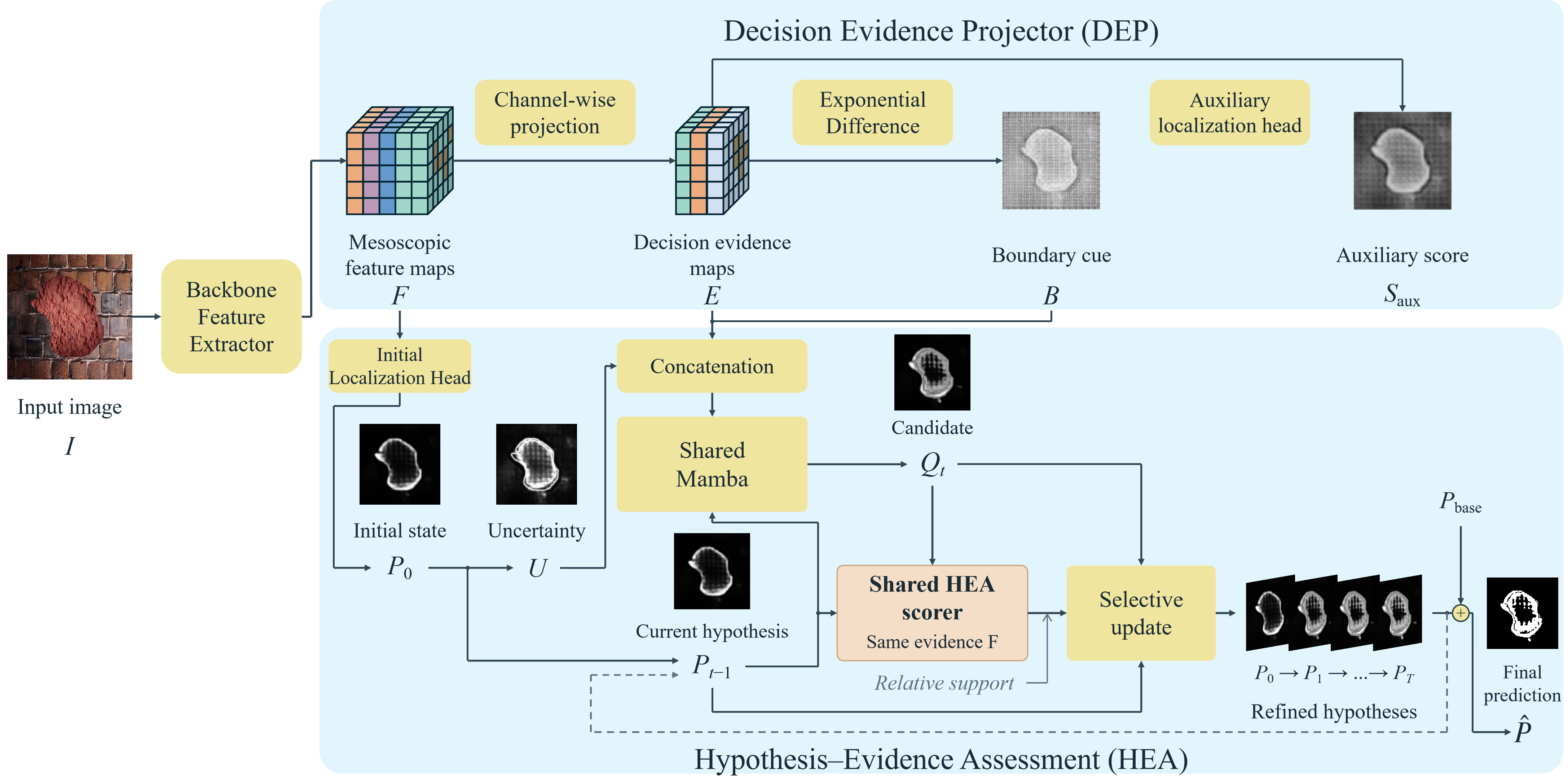}
\caption{Hypothesis--evidence interaction for progressive localization.
DEP projects mesoscopic features for a shared Mamba candidate generator.
HEA scores the candidate and current hypotheses against the same evidence $F$; their relative support controls selective updating.
The updated hypothesis conditions the next iteration, and the final prediction combines the refined map with the backbone output.}
\label{fig:architecture}
\end{figure*}

As illustrated in Fig.~\ref{fig:architecture}, our framework couples a Mamba-based candidate generator with a shared Hypothesis--Evidence Assessment (HEA) module.
Starting from the backbone's mesoscopic features and initial localization map, the generator explores contextual corrections, while HEA determines their relative support by assessing both hypotheses against the same evidence.
Using the current hypothesis as a reference makes assessment specific to the proposed change, and the resulting comparison guides selective updates across the image.
The revised hypothesis then informs subsequent proposals and assessments, forming an iterative interaction between hypothesis revision and evidence evaluation.
This coupling allows the model to explore corrections while preserving the current prediction wherever the candidate lacks positive relative support.

\subsection{Evidence-Guided Candidate Generation}
\label{sec:candidate}

Given an input image, a pretrained backbone produces mesoscopic features $F$ and an initial localization map $P_0$ on a common spatial grid, which serves as the working resolution for iterative updating.
The backbone remains frozen while the proposed modules are trained.
Candidate generation combines the current localization hypothesis with spatial context from $F$, so that a proposal can address errors remaining in the current state.
A lightweight decision evidence projector (DEP) maps $F$ to a compact representation $E=\mathrm{DEP}(F)$ using a $1\times1$ convolution, batch normalization, and GELU.
This projection provides a compact input for contextual modeling, while the original features $F$ remain available to the assessment module.
To indicate ambiguity in the initial prediction, we compute an uncertainty cue $U$ as the pixelwise Shannon entropy of the Bernoulli distribution defined by $P_0$.
We also derive a boundary cue from spatial variations in the projected features:
\begin{equation}
 B=\exp\!\left(-\kappa\,\operatorname{mean}_{c}
       (|\nabla_x E|+|\nabla_y E|)\right),
 \label{eq:boundary}
\end{equation}
where $\nabla_x$ and $\nabla_y$ denote horizontal and vertical feature differences, $\operatorname{mean}_{c}$ averages over channels, and $\kappa>0$ controls how rapidly the cue decays with increasing gradient magnitude.
Thus, $B$ takes lower values where the projected features change sharply.
Both cues remain fixed across iterations.

At step $t$, the generator combines these cues with the current hypothesis:
\begin{equation}
 Q_t=\mathcal{M}\!\left([E,P_{t-1},U,B]\right).
 \label{eq:candidate}
\end{equation}
Here, brackets denote channel concatenation, and $\mathcal{M}$ flattens the spatial grid into a sequence, adds learned positional embeddings, and applies Mamba~\cite{gu2024mamba} followed by a linear projection and sigmoid.
Mamba allows each proposal to incorporate context beyond the local neighborhood.
Reusing the generator across iterations lets the same correction mechanism respond to the changing hypothesis.
The resulting $Q_t$ is passed to the assessment module before updating the state.

\subsection{Hypothesis--Evidence Assessment}
\label{sec:scorer}

\paragraph{Relative evidence support.}
HEA assesses a proposed revision by comparing the evidence support for the candidate $Q_t$ with that for the current hypothesis $P_{t-1}$:
\begin{equation}
 \Delta_t=S(F,Q_t)-S(F,P_{t-1}),
 \label{eq:delta}
\end{equation}
where $S$ is a learnable scoring function that evaluates a hypothesis against the forensic evidence $F$ and produces a pixelwise support map.
Both hypotheses are evaluated using the same network and parameters, placing their scores on a common learned scale.
Using the current hypothesis as a reference makes $\Delta_t$ specific to the proposed revision, including corrections to initially confident but incorrect predictions.
The meaning of relative support is learned from the comparative supervision in Sec.~\ref{sec:loss}.
A positive difference favors the candidate, while a nonpositive difference provides no support for adopting the revision.

To compute $S$, we first encode the mesoscopic features as $V=\phi(F)$ and then jointly process the encoded evidence and a hypothesis map $H$:
\begin{equation}
 S(F,H)=\psi([V,H]),
 \label{eq:score}
\end{equation}
where $H$ denotes either $Q_t$ or $P_{t-1}$.
The evidence encoder $\phi$ is a $3\times3$ convolution with GELU, and the compatibility network $\psi$ combines $1\times1$ and $3\times3$ convolutions with nonlinear activations.
Processing $V$ and $H$ jointly allows the support assigned to a hypothesis to depend on its compatibility with the forensic evidence.

\paragraph{Selective hypothesis updating.}
Relative evidence support controls how much of the proposed revision is adopted:
\begin{align}
 A_t&=\left[2\sigma(\gamma\Delta_t)-1\right]_+,
 \label{eq:strength}\\
 P_t&=P_{t-1}+A_t\odot(Q_t-P_{t-1}),
 \label{eq:update}
\end{align}
where $\sigma$ is the sigmoid function, $[x]_+=\max(x,0)$, $\odot$ denotes elementwise multiplication, and $\gamma>0$ controls the sensitivity of the update strength to relative support.
A larger $\gamma$ yields a stronger update for the same positive score difference, while nonpositive support leaves the current prediction unchanged.
The current hypothesis thus provides a fallback when the assessor does not favor a revision, and the continuous strength allows partial adoption when support is limited.

During training, we fuse the refined prediction with the frozen backbone output to provide a fixed reference for optimization, while deep supervision directly trains the candidate maps.
This fusion serves as a training aid. At inference, we use only the upsampled refined prediction.
The fused prediction is
$\widehat P=\alpha\mathcal{U}(P_T)+(1-\alpha)P_{\mathrm{base}}$,
where $P_{\mathrm{base}}$ is the backbone's full-resolution probability map, $\mathcal{U}$ upsamples the refined hypothesis to the input resolution, and the learnable weight $\alpha\in[0,1]$ balances the two predictions.
The initial map $P_0$ is obtained by applying sigmoid directly to the working-resolution backbone logits, whereas $P_{\mathrm{base}}$ is obtained by upsampling the same logits before applying sigmoid.

\subsection{Training Objectives}
\label{sec:loss}

Our objective trains the generator to produce accurate localization candidates and the scorer to identify beneficial revisions.

\paragraph{Comparative supervision.}
We supervise relative evidence support using the improvement a candidate makes over the current hypothesis.
For a binary ground-truth label $y_i$ at the working resolution, define
\begin{equation}
 d_{t,i}=|P_{t-1,i}-y_i|-|Q_{t,i}-y_i|,
 \qquad b_{t,i}=\mathbf{1}[d_{t,i}>0].
 \label{eq:improvement}
\end{equation}
Thus, $b_{t,i}$ indicates whether the candidate is closer to the ground truth.
These targets are constructed from the model's current predictions and the existing pixel annotations, providing direct supervision for candidate adoption at each iteration.
We train the scorer to predict this preference through
\begin{equation}
 \mathcal{L}_{\mathrm{assess}}
 =\frac{1}{T}\sum_{t=1}^{T}
 \mathcal{L}_{\mathrm{gBCE}}(\sigma(\gamma\Delta_t),b_t).
 \label{eq:assessloss}
\end{equation}
Here, $\mathcal{L}_{\mathrm{gBCE}}$ partitions valid pixels into four groups by crossing the ground-truth foreground/background labels with the beneficial/harmful revision labels, excluding negligible revisions.
BCE is averaged within each nonempty group and then equally across these groups.
This balancing prevents abundant background pixels or one revision category from dominating assessment learning.

\paragraph{Localization supervision.}
We directly supervise each candidate so that proposal generation learns to correct localization errors even when a revision is not adopted.
The candidate loss combines pixelwise and region-level supervision:
\begin{equation}
 \mathcal{L}_{\mathrm{deep}}
 =\sum_{t=1}^{T}w_t
 \left[\mathcal{L}_{\mathrm{4BCE}}(Q_t^\uparrow,Y)
       +\mathcal{L}_{\mathrm{Dice}}(Q_t^\uparrow,Y)\right],
 \label{eq:deep}
\end{equation}
where $Y$ is the full-resolution ground-truth mask, $Q_t^\uparrow$ is the upsampled candidate, and the increasing temporal weight $w_t$ emphasizes later predictions.
The group-balanced loss $\mathcal{L}_{\mathrm{4BCE}}$ gives equal weight to the nonempty TP/FN/FP/TN groups defined by the binarized backbone prediction $P_{\mathrm{base}}$ and $Y$, while Dice loss encourages region overlap.
An auxiliary head on $E$ receives class-balanced BCE supervision $\mathcal{L}_{\mathrm{aux}}$ to encourage the projected features to preserve localization information.

The final prediction receives BCE supervision, giving the overall objective
\begin{equation}
 \mathcal{L}
 =\mathcal{L}_{\mathrm{BCE}}(\widehat P,Y)
 +\lambda_{\mathrm{deep}}\mathcal{L}_{\mathrm{deep}}
 +\lambda_{\mathrm{aux}}\mathcal{L}_{\mathrm{aux}}
 +\lambda_{\mathrm{assess}}\mathcal{L}_{\mathrm{assess}}.
 \label{eq:total}
\end{equation}

The weights $\lambda_{\mathrm{deep}}$, $\lambda_{\mathrm{aux}}$, and $\lambda_{\mathrm{assess}}$ balance candidate supervision, projected-feature supervision, and comparative assessment against the final localization objective.
The scorer is optimized only through $\mathcal{L}_{\mathrm{assess}}$, while localization losses train the generator and evidence projector.
Preference targets are recomputed along the unrolled updates, allowing the feedback process to be trained directly with pixel annotations.
Further analysis of the supervision and update rule is provided in Appendix~\ref{sec:progressive_learning}.

\section{Experiments}
\label{sec:experiments}

\subsection{Datasets}
\label{sec:datasets}

We follow the Protocol-CAT training setup~\cite{ma2024imdl} provided by IMDL-BenCo, combining CASIAv2~\cite{dong2013casia}, Fantastic Reality~\cite{DBLP:conf/nips/KniazKR19}, IMD2020~\cite{DBLP:conf/wacv/NovozamskyMS20}, and tampered COCO~\cite{DBLP:journals/ijcv/KwonNYLK22}.
For conventional manipulation evaluation, we use NIST16~\cite{DBLP:conf/wacv/GuanK0LYDZKSF19} and CASIAv1~\cite{dong2013casia}.
For AI-generated manipulations, we use AutoSplice~\cite{jia2023autosplice} (3,621 images) and a sampled subset of SAGI-D~\cite{giakoumoglou2025sagi}, denoted SAGI-D-9K.
The latter provides generative inpainting examples and separate training and test splits.

\subsection{Experimental Protocols}
\label{sec:exp_protocols}

We evaluate the proposed method under two protocols.
\textbf{Protocol I} is used for state-of-the-art comparison.
The training setup follows the standardized Protocol-CAT setting~\cite{ma2024imdl}.
The model is evaluated on NIST16, CASIAv1, AutoSplice, and SAGI-D-9K.
We study the open-ended challenge by training on conventional manipulations and evaluating on AI-generated manipulations from AutoSplice and SAGI-D-9K, neither of which is used for gradient-based training under this protocol.

\textbf{Protocol II} is used to study the effect of introducing AI-generated manipulation data during training.
Based on Protocol I, we further use the training split of SAGI-D-9K for training, and evaluate on the SAGI-D-9K test split and AutoSplice.
This protocol examines how expanding the training coverage affects localization performance on AI-generated manipulations.

\subsection{Implementation Details}
\label{sec:impl}

Unless otherwise specified, we follow the standardized Protocol-CAT setting.
We use Mesorch~\cite{zhu2025mesoscopic} to extract mesoscopic features.
The CNN~\cite{DBLP:conf/cvpr/0003MWFDX22} and Transformer~\cite{DBLP:conf/nips/XieWYAAL21} branches receive RGB concatenated with high- and low-frequency components, respectively.
Four aligned response maps from each branch form the eight-channel feature map $F$, combining local forensic cues with global context.

Training proceeds in two stages.
We first pretrain the backbone under the corresponding protocol, then freeze it while training DEP, the candidate generator, and HEA.
Protocol I uses the Protocol-CAT training set, while Protocol II additionally incorporates the SAGI-D-9K training split.

Optimization settings, hyperparameters, and loss weights are provided in Appendix~\ref{sec:supp_implementation}.

\subsection{Comparison with State-of-the-Art Methods}

\begin{table*}[t]
\centering
\setlength{\tabcolsep}{4pt}
\renewcommand{\arraystretch}{1.15}
\caption{Comparison under Protocol I.
Each entry is reported as \textbf{F1-score / Permute F1-score}.
-- denotes unreported results.
DiffIML$^\dagger$ additionally uses compRAISE for training and horizontal-flip test-time augmentation, so its setting differs from ours.
The best and second-best reported results are bold and underlined, respectively.}
\label{tab:sota_protocol1}
\resizebox{\textwidth}{!}{%
\begin{tabular}{lcccccc}
\toprule
\multirow{2}{*}{Method} & \multicolumn{6}{c}{F1-score / Permute F1-score} \\
\cmidrule(lr){2-7}
& NIST16 & CASIAv1 & AutoSplice & SAGI-D-9K
& Traditional Avg. & AIGC Avg. \\
\midrule

MVSS-Net
& 0.2876 / 0.3377
& 0.5486 / 0.5707
& 0.3779 / 0.5570
& -- / --
& 0.4181 / 0.4542
& -- / -- \\

PSCC-Net
& 0.3681 / 0.4143
& 0.5472 / 0.5583
& 0.5531 / 0.6831
& -- / --
& 0.4577 / 0.4863
& -- / -- \\

IML-ViT
& 0.1100 / 0.1503
& 0.7392 / 0.7512
& 0.2687 / 0.5676
& -- / --
& 0.4246 / 0.4508
& -- / -- \\

TruFor
& 0.3480 / 0.4046
& 0.8176 / 0.8340
& 0.3830 / \underline{0.6910}
& -- / --
& 0.5828 / 0.6193
& -- / -- \\

Co-Transformers
& \underline{0.4551} / \underline{0.4994}
& \textbf{0.8548} / \textbf{0.8642}
& 0.5360 / 0.6906
& -- / --
& \textbf{0.6550} / \textbf{0.6818}
& -- / -- \\

Mesorch
& 0.3921 / 0.4514
& \underline{0.8398} / \underline{0.8472}
& 0.4004 / 0.6709
& \underline{0.1919} / \underline{0.4437}
& 0.6160 / 0.6493
& \underline{0.2962} / \underline{0.5573} \\

DiffIML$^\dagger$
& 0.3410 / --
& 0.8280 / --
& \underline{0.6120} / --
& -- / --
& 0.5845 / --
& -- / -- \\

Ours
& \textbf{0.4689} / \textbf{0.5011}
& 0.8172 / 0.8206
& \textbf{0.6353} / \textbf{0.7077}
& \textbf{0.3228} / \textbf{0.4872}
& \underline{0.6431} / \underline{0.6609}
& \textbf{0.4791} / \textbf{0.5974} \\

\bottomrule
\end{tabular}%
}
\end{table*}

Table~\ref{tab:sota_protocol1} compares our method with representative forgery localization methods under Protocol I, using conventional manipulations for gradient training and evaluating on both conventional and AI-generated benchmarks.
The comparisons cover artifact-based, correlation-based, and Transformer-based localization, with Mesorch~\cite{zhu2025mesoscopic} serving as our direct backbone baseline.
For MVSS-Net~\cite{DBLP:journals/pami/DongCHCL23}, PSCC-Net~\cite{DBLP:journals/tcsv/LiuLCL22}, IML-ViT~\cite{ma2023iml}, TruFor~\cite{DBLP:conf/cvpr/GuillaroCSDV23}, Mesorch, and Co-Transformers~\cite{DBLP:conf/aaai/ZhangFWKYN26}, the NIST16, CASIAv1, and AutoSplice results are taken from Table~1 of Zhang et al.
DiffIML results are taken from its original paper~\cite{wang2026diffiml}.
We obtain the Mesorch results on SAGI-D-9K through our own evaluation using the same protocol as our method.

On conventional benchmarks, our method improves the Mesorch backbone's NIST16 F1 from 0.3921 to 0.4689 and its traditional average F1 from 0.6160 to 0.6431.
On CASIAv1, however, Mesorch already provides strong localization, with an F1 of 0.8398 and macro-averaged pixel recall of 87.34\%.
The high scores of several strong baselines suggest that CASIAv1 is relatively less challenging for current methods and less informative about localization beyond conventional training coverage.
At the fixed threshold of 0.5, our method increases recall to 89.67\%, while precision drops from 86.94\% to 79.97\%.
The high initial recall leaves less room to recover missed manipulated pixels, while revisions that include background pixels reduce precision.
On this benchmark, the precision cost outweighs the gain in coverage, and F1 decreases to 0.8172.
Co-Transformers' stronger CASIAv1 performance does not carry over to AutoSplice, where our method performs better, consistent with our focus on manipulation patterns beyond conventional training coverage.
Our advantage over DiffIML on AutoSplice, despite its additional compRAISE training data and test-time augmentation, further supports the value of hypothesis--evidence interaction in this open-ended setting.

\begin{figure*}[!t]
\centering
\includegraphics[width=0.95\textwidth]{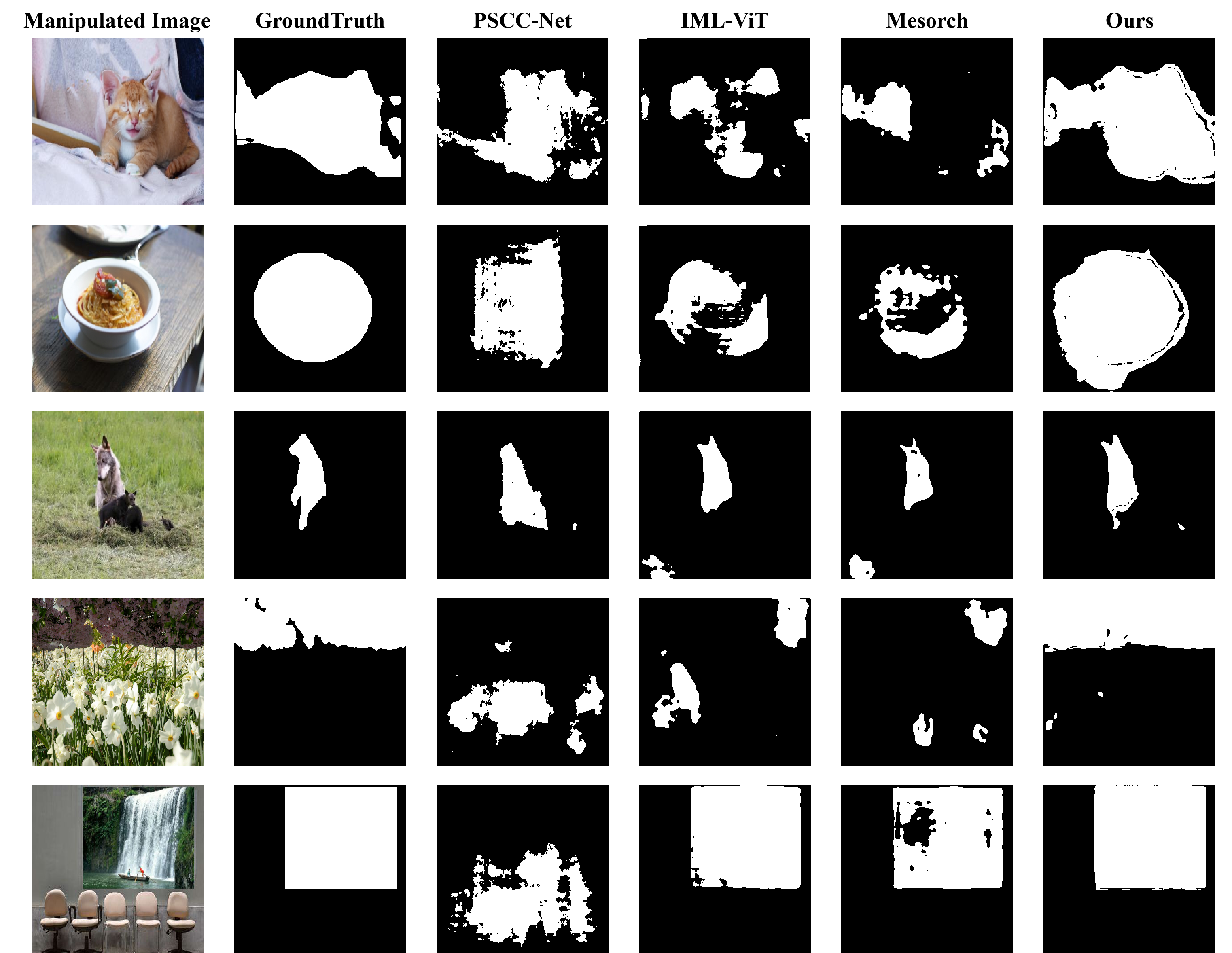}
\caption{Qualitative comparison with PSCC-Net, IML-ViT, and Mesorch.
Columns show the manipulated image, ground truth, the three baseline predictions, and our prediction.
The examples include recovered regions, incomplete masks, and residual false positives.}
\label{fig:vissota}
\end{figure*}

Our method primarily targets open-ended forgery localization, where test manipulation patterns may extend beyond the training coverage.
On AutoSplice and SAGI-D-9K, Mesorch's initial F1 scores are 0.4004 and 0.1919; our method improves them to 0.6353 and 0.3228, gains of 23.49 and 13.09 percentage points, respectively.
Among methods reporting both AI-generated benchmarks, the AIGC average F1 increases from Mesorch's 0.2962 to 0.4791.
Under Protocol I, the proposed modules learn from conventional manipulations without using AutoSplice or SAGI-D-9K for gradient-based training.
These observations are consistent with our expectation that hypothesis--evidence interaction can improve the use of existing mesoscopic evidence when manipulation patterns differ from those seen during training.
With the backbone frozen, the gains on AI-generated benchmarks support the value of revising localization hypotheses against the same forensic evidence.
Figure~\ref{fig:vissota} provides representative qualitative comparisons.

\subsection{Effect of Introducing AI-Generated Training Data}
\label{sec:protocol2}

\begin{table}[t]
\centering
\caption{Comparison under Protocol II with the SAGI-D-9K training split added to Protocol-CAT.
Each entry is reported as F1-score/ Permute F1-score.
The best results are highlighted in bold.}
\label{tab:protocol2_aigc_training}
\begin{tabular*}{\columnwidth}{@{\extracolsep{\fill}}lcc}
\toprule
\multirow{2}{*}{Method}
& \multicolumn{2}{c}{F1 / Permute F1} \\
\cmidrule(lr){2-3}
& AutoSplice
& SAGI-D-9K \\
\midrule
PSCC-Net
& 0.5501 / 0.5508
& 0.3523 / 0.3549 \\
IML-ViT
& 0.5563 / 0.6293
& 0.5810 / 0.6348 \\
Mesorch
& 0.5319 / 0.7174
& 0.7399 / 0.7531 \\
Ours
& \textbf{0.6030} / \textbf{0.7333}
& \textbf{0.7418} / \textbf{0.7555} \\
\bottomrule
\end{tabular*}
\end{table}

Table~\ref{tab:protocol2_aigc_training} compares models after adding the SAGI-D-9K training split, with AutoSplice excluded from gradient training.
Our method achieves the highest F1 and Permute F1 on both benchmarks.
Its F1 gains over Mesorch are 7.11 and 0.19 percentage points on AutoSplice and SAGI-D-9K, respectively.

Comparing Protocols I and II reveals different effects on backbone localization and subsequent refinement.
Under Protocol II, Mesorch improves on both datasets, substantially narrowing our additional gain on SAGI-D-9K, while our AutoSplice F1 declines.
SAGI-D covers diverse generative forgery patterns, but its inpainting pipelines still differ from the DALL-E 2 manipulations in AutoSplice.
Learning the added patterns enables the backbone to address more localization errors, leaving less room for subsequent correction.
Meanwhile, candidate generation is conditioned on the backbone's predictions, and HEA learns whether each revision improves them.
The resulting correction behavior may therefore adapt to the added forgery patterns, with smaller benefits when transferred to different manipulations.
AutoSplice Permute F1 nevertheless increases, suggesting that the ordinary-F1 decline may partly involve foreground/background assignment.

\begin{figure*}[!t]
\centering
\includegraphics[width=0.95\textwidth]{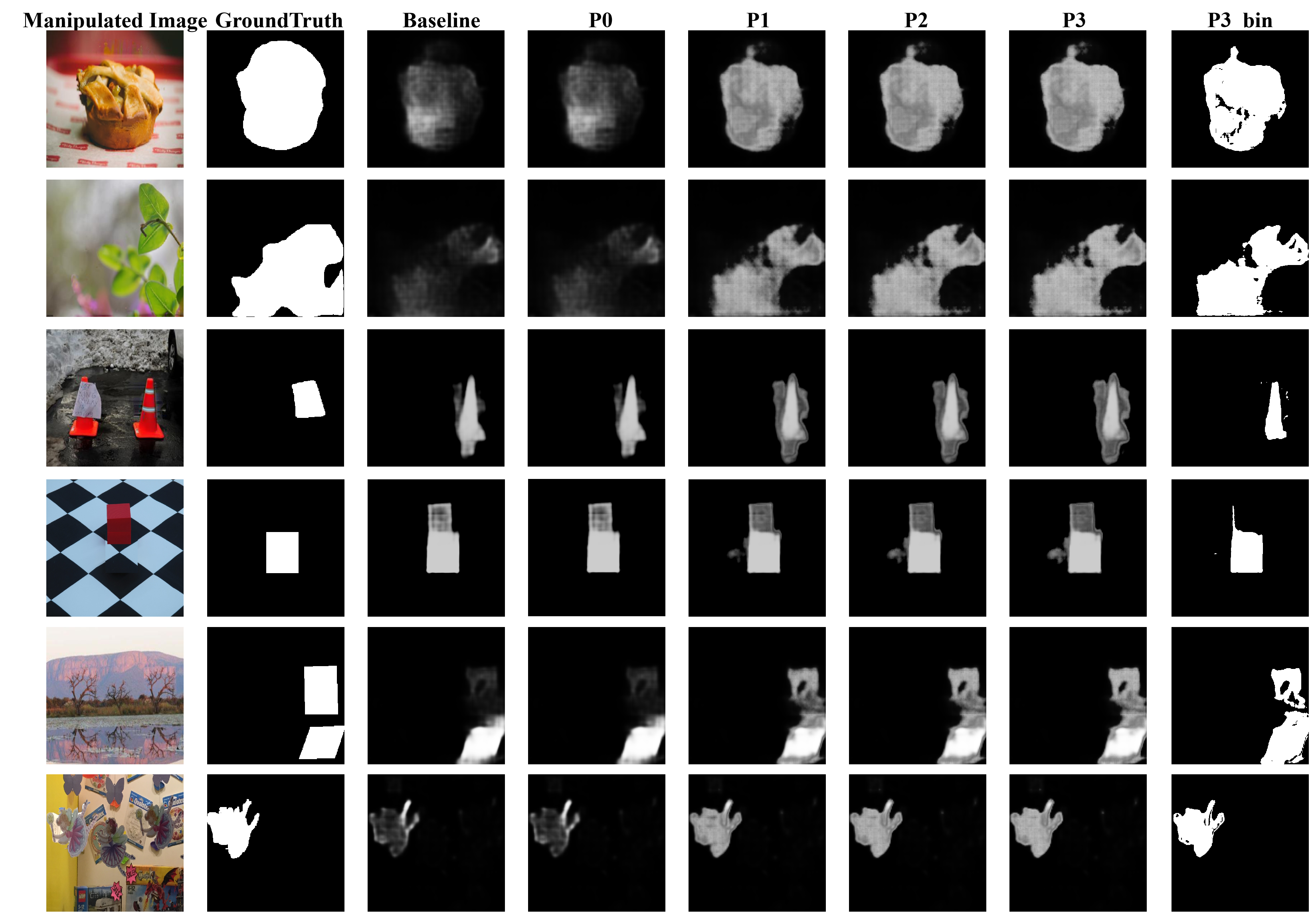}
\caption{Progressive localization with $T=3$.
Columns show the input image, ground truth, backbone prediction, the hypotheses $P_0$--$P_3$, and the binarized $P_3$.}
\label{fig:visp}
\end{figure*}

\subsection{Ablation Study}
\label{sec:ablation}
\label{sec:overall_ablation}

\begin{table}[t]
\centering
\caption{Component and objective ablations evaluated on AutoSplice under Protocol~II. The best scores are bold.}
\label{tab:ablation}
\begin{tabular*}{\columnwidth}{@{\extracolsep{\fill}}lcc}
\toprule
Variant & F1 & Permute F1 \\
\midrule
Baseline (Mesorch) & 0.5319 & 0.7174 \\
w/o DEP & 0.5365 & 0.7165 \\
w/o uncertainty $U$ & 0.6580 & 0.7472 \\
w/o boundary cue $B$ & 0.6760 & 0.7592 \\
Probability-only HEA & 0.6500 & 0.7519 \\
Direct update (w/o HEA) & 0.6487 & 0.7138 \\
w/o $\mathcal{L}_{\mathrm{aux}}$ & 0.6780 & 0.7573 \\
w/o $\mathcal{L}_{\mathrm{deep}}$ & 0.5319 & 0.7174 \\
\textbf{Full model} & \textbf{0.6968} & \textbf{0.7664} \\
\bottomrule
\end{tabular*}
\end{table}

Table~\ref{tab:ablation} isolates the contributions of the proposed modules.
Figure~\ref{fig:visp} illustrates the progressive localization maps from $P_0$ to $P_3$.
Removing DEP produces the largest component drop, reducing F1 from 0.6968 to 0.5365; this variant also omits its auxiliary supervision.
Removing $U$ or $B$ reduces F1 to 0.6580 or 0.6760.
Probability-only assessment yields 0.6500 F1, supporting the contribution of image-dependent evidence to the scorer.
Direct candidate replacement gives 0.6487 F1, 4.81 points below selective updating.
This variant also removes $\mathcal{L}_{\mathrm{assess}}$ together with HEA, so we do not list a separate assessment-loss ablation.

Removing auxiliary supervision while retaining DEP yields 0.6780 F1 and 0.7573 Permute F1, decreasing the two scores by 1.88 and 0.91 percentage points, respectively.
This supports the role of $\mathcal{L}_{\mathrm{aux}}$ in preserving localization information in the projected features.
Without candidate deep supervision, performance matches the backbone at the reported precision.
This loss supplies a learning signal even for candidate revisions rejected by HEA.
Parameter studies in Appendix~\ref{sec:supp_parameters} favor $T=3$ and $\gamma=1$ by validation F1.
The deep-supervision sweep peaks at $\lambda_{\mathrm{deep}}=0.4$ (0.7003 F1); the ablation reference retains 1.0 (0.6968).

\subsection{Robustness to Input Perturbations}
\label{sec:robustness}

\begin{table}[t]
\centering
\caption{Robustness under Protocol~II: average F1 over seven conditions, including clean input. Full results are in Table~\ref{tab:robustness}.}
\label{tab:robustness_summary}
\begin{tabular*}{\columnwidth}{@{\extracolsep{\fill}}lccc}
\toprule
Dataset & Perturbation & Mesorch & Ours \\
\midrule
\multirow{3}{*}{AutoSplice} & Noise & 0.2905 & \textbf{0.3691} \\
 & Blur & 0.3619 & \textbf{0.4541} \\
 & JPEG & 0.3420 & \textbf{0.4254} \\
\midrule
\multirow{3}{*}{SAGI-D-9K} & Noise & 0.7151 & \textbf{0.7223} \\
 & Blur & 0.6802 & \textbf{0.6839} \\
 & JPEG & 0.7001 & \textbf{0.7073} \\
\bottomrule
\end{tabular*}
\end{table}

We test Gaussian noise, Gaussian blur, and JPEG compression under Protocol II.
This evaluation uses the same checkpoint as Table~\ref{tab:protocol2_aigc_training}.
Table~\ref{tab:robustness_summary} averages seven conditions per perturbation, including clean input; full results are in Appendix~\ref{sec:supp_robustness}.
On AutoSplice, our average F1 exceeds Mesorch by 7.86, 9.22, and 8.34 percentage points for noise, blur, and JPEG compression.
On SAGI-D-9K, the respective gains are 0.72, 0.37, and 0.72 points.
The sole pointwise exception is SAGI-D-9K with blur kernel 11 (0.6853 versus 0.6856).
Thus, all six averages improve, although not every individual condition does.

\subsection{Qualitative Analysis}
\label{sec:qualitative}

Figure~\ref{fig:vissota} compares the updated localization results with representative baselines, while Fig.~\ref{fig:visp} shows the progressive prediction maps in the main paper.
The displayed examples include incomplete masks, residual false positives, and changes that do not improve overlap.
Most visible changes in the progressive examples occur between $P_0$ and $P_1$, with smaller differences in later steps.
These examples illustrate the update behavior without implying that every revision improves localization.

\section{Conclusion}

We presented a hypothesis--evidence interaction framework for localizing open-ended image forgeries.
The shared HEA module compares each proposed change by comparing current and candidate hypotheses against the same forensic evidence.
This lets the generator explore contextual corrections while relative support determines where and how strongly they are adopted, preserving current predictions where support is nonpositive.
Accepted revisions inform subsequent generation and assessment, allowing the same evidence to be reconsidered as the hypothesis evolves.
Experiments on conventional and AI-generated benchmarks demonstrate the effectiveness of this interaction, particularly on AI-generated forgeries under conventional-only training.
Although refinement tends to stabilize within a few iterations, repeated generation and assessment incur computational overhead.
Future work will explore adaptive selection of the number of update steps to balance accuracy and efficiency for each image.

\section*{LLM Usage Statement}
OpenAI ChatGPT was used to assist with language editing, grammar correction and LaTeX formatting and compilation.
Its use included suggesting wording, reorganizing explanations, and improving the presentation of existing experimental results.
The authors are responsible for checking these revisions and verifying all scientific claims, experimental results, and the final manuscript.
{
    \small
    \bibliographystyle{ieeenat_fullname}
    \bibliography{main}
}

\clearpage
\appendix
\setcounter{page}{1}
\renewcommand{\thepage}{S\arabic{page}}
\setcounter{figure}{0}
\setcounter{table}{0}
\renewcommand{\theHfigure}{supp.\arabic{figure}}
\renewcommand{\theHtable}{supp.\arabic{table}}
\renewcommand{\thefigure}{S\arabic{figure}}
\renewcommand{\thetable}{S\arabic{table}}
\numberwithin{equation}{section}
\maketitlesupplementary

\section{Supervised Learning of Progressive Updates}
\label{sec:progressive_learning}

The assessment objective and update rule are linked by the change in localization error.
For a binary label $y_i$ and probability-valued hypotheses, absolute error is affine on $[0,1]$.
Combining \cref{eq:update,eq:improvement} therefore gives
\begin{equation}
 |P_{t-1,i}-y_i|-|P_{t,i}-y_i|
 = A_{t,i}\,d_{t,i}.
 \label{eq:accepted_improvement}
\end{equation}
Equation~\eqref{eq:accepted_improvement} expresses the reduction in pixelwise probability error as the candidate's improvement $d_{t,i}$ scaled by the applied update strength $A_{t,i}$.
The assessment loss trains HEA to identify beneficial revisions, while localization supervision trains the generator to propose them.
When HEA favors a candidate, the update strength determines how much of the proposed change is applied.
The updated hypothesis then conditions both candidate generation and assessment in the next step.
Once selective updating is enabled during training, we unroll this recurrence and recompute the preference targets for the current and candidate hypotheses at each step.
This lets HEA learn from the revisions encountered along the refinement trajectory.
At inference, the same recurrence uses the learned scorer to guide each update.

Our formulation differs from reinforcement learning in how feedback is used to train update decisions.
Reinforcement learning uses rewards to optimize an update policy, as exemplified by CoDE~\cite{peng2024code}, which formulates iterative localization as a Markov decision process with sampled continuous actions.
In our setting, pixel annotations reveal whether a proposed revision reduces the error of the current hypothesis at each pixel.
This gives us a direct target for training HEA to assess each candidate revision.

\section{Additional Implementation Details}
\label{sec:supp_implementation}

\paragraph{Feature and module dimensions.}
Four aligned, single-channel response maps from each Mesorch~\cite{zhu2025mesoscopic} branch form the eight-channel evidence $F$.
DEP projects $F$ to four channels for candidate generation.
The Mamba module uses model dimension 7, state dimension 16, convolution width 4, and expansion factor 2; the scorer has hidden width 32.

\paragraph{Optimization.}
Following the two-stage setup in Sec.~\ref{sec:impl}, we freeze the pretrained backbone and train the proposed modules for 10 epochs.
Images are resized to $512\times512$. We use AdamW with weight decay 0.05.
We train on four GPUs with a batch size of 8 per GPU (effective batch
size 32), with no gradient accumulation.
The scorer and fusion parameter use a peak learning rate of $5\times10^{-5}$ and
a scheduled minimum of $5\times10^{-7}$. The candidate branch, including
DEP, uses a peak learning rate of $5\times10^{-4}$ and a minimum of $5\times10^{-6}$. Both groups use two epochs
of learning-rate warmup.

\paragraph{Losses and update schedule.}
We set $(\lambda_{\mathrm{deep}},\lambda_{\mathrm{aux}},
\lambda_{\mathrm{assess}})=(1,0.01,0.1)$.
The group-balanced BCE in \cref{eq:assessloss} uses only pixels with $|d_{t,i}|>10^{-3}$; a step without valid pixels contributes zero assessment loss.
We detach the
scorer inputs $(F,P_{t-1},Q_t)$ and block localization gradients through
the applied update strength $A_t$. The scorer is trained from the first
epoch. During the first three training epochs, we use direct candidate
updates $P_t=Q_t$; subsequent epochs use the selective update in
\cref{eq:update}.

\paragraph{Refinement and inference.}
Unless otherwise specified, we use $T=3$ refinement steps during
training and inference in both protocols. We set $\kappa=2$ for the
boundary cue and $\gamma=1$ for relative support scaling.
We learn the fusion weight $\alpha$ during training
and set $\alpha=1$ at inference, so the final prediction uses only the
refined mask.

\paragraph{Evaluation metrics.}
We report foreground F1 at a threshold of 0.5 and Permute F1, which selects the better of the original and inverted binary mask per image.
The traditional average covers NIST16 and CASIAv1; the AIGC average covers AutoSplice and SAGI-D-9K.
An average is reported only when all corresponding dataset results are available.

\section{Additional Parameter Studies}
\label{sec:supp_parameters}

We vary one parameter at a time on AutoSplice under Protocol~II, fixing the others at $(\gamma,T,\lambda_{\mathrm{deep}})=(1,3,1)$.
Each configuration is trained separately; F1 and Permute F1 are reported from the same checkpoint selected by validation F1.

\begin{figure*}[!t]
\centering
\begin{tikzpicture}
\definecolor{sensblue}{RGB}{43,107,176}
\definecolor{sensred}{RGB}{210,91,79}
\pgfplotsset{senscommon/.style={
 width=0.20\textwidth,height=0.17\textwidth,scale only axis,
 axis line style={black!60,line width=0.5pt},tick style={line width=0.4pt},
 tick align=inside,tick label style={font=\scriptsize},label style={font=\small},
 title style={font=\small,yshift=4pt},scaled ticks=false,
 yticklabel style={/pgf/number format/fixed,/pgf/number format/precision=2},
 every axis plot/.append style={line width=1.1pt,mark size=2pt},enlargelimits=false,
 },sensleft/.style={senscommon,axis y line*=left,axis x line*=bottom,
 ymin=0.52,ymax=0.72,ytick={0.55,0.60,0.65,0.70},
 yticklabel style={text=sensblue},ylabel style={text=sensblue},
 ymajorgrids=true,grid style={black!12,densely dotted,line width=0.35pt}},
 sensright/.style={senscommon,axis y line*=right,axis x line=none,
 ymin=0.71,ymax=0.78,ytick={0.72,0.74,0.76,0.78},
 yticklabel style={text=sensred},ylabel style={text=sensred}}}
\begin{groupplot}[sensleft,group style={group size=3 by 1,horizontal sep=0.11\textwidth}]
\nextgroupplot[title={(a) HEA scaling},xmin=-0.15,xmax=3.15,xtick={0,1,2,3},xticklabels={0.5,1,2,4},
 ymin=0.66,ymax=0.705,ytick={0.66,0.67,0.68,0.69,0.70},
 xlabel={$\gamma$},ylabel={F1 $\uparrow$},legend to name=sensitivitylegend,
 legend style={font=\scriptsize,draw=none,fill=none,legend columns=3,
 /tikz/every even column/.append style={column sep=6pt}}]
\addplot[sensblue,mark=*] coordinates {(0,0.6966)(1,0.6968)(2,0.6785)(3,0.6655)};
\addlegendentry{F1 (left)}
\addlegendimage{sensred,mark=square*}
\addlegendentry{Permute F1 (right)}
\addlegendimage{black!70,only marks,mark=star,mark size=3pt}
\addlegendentry{Main setting}
\addplot[sensblue,only marks,mark=star,mark size=3.6pt,forget plot] coordinates {(1,0.6968)};
\nextgroupplot[title={(b) Refinement steps},xmin=0.85,xmax=4.15,xtick={1,2,3,4},xlabel={$T$},
 ymin=0.684,ymax=0.700,ytick={0.684,0.688,0.692,0.696,0.700},
 yticklabel style={text=sensblue,/pgf/number format/fixed,/pgf/number format/precision=3}]
\addplot[sensblue,mark=*] coordinates {(1,0.6862)(2,0.6904)(3,0.6968)(4,0.6947)};
\addplot[sensblue,only marks,mark=star,mark size=3.6pt] coordinates {(3,0.6968)};
\nextgroupplot[title={(c) Deep supervision},xmin=-0.05,xmax=1.25,xtick={0,0.2,0.4,0.6,0.8,1,1.2},xlabel={$\lambda_{\mathrm{deep}}$}]
\addplot[sensblue,mark=*] coordinates {(0,0.5319)(0.2,0.6869)(0.4,0.7003)(0.6,0.6962)(0.8,0.6984)(1,0.6968)(1.2,0.6966)};
\addplot[sensblue,only marks,mark=star,mark size=3.6pt] coordinates {(1,0.6968)};
\end{groupplot}
\begin{axis}[sensright,at={(group c1r1.south west)},anchor=south west,xmin=-0.15,xmax=3.15,
 ymin=0.74,ymax=0.775,ytick={0.74,0.75,0.76,0.77}]
\addplot[sensred,mark=square*] coordinates {(0,0.7679)(1,0.7664)(2,0.7506)(3,0.7436)};
\addplot[sensred,only marks,mark=star,mark size=3.6pt] coordinates {(1,0.7664)};
\end{axis}
\begin{axis}[sensright,at={(group c2r1.south west)},anchor=south west,xmin=0.85,xmax=4.15,
 ymin=0.760,ymax=0.770,ytick={0.760,0.762,0.764,0.766,0.768,0.770},
 yticklabel style={text=sensred,/pgf/number format/fixed,/pgf/number format/precision=3}]
\addplot[sensred,mark=square*] coordinates {(1,0.7682)(2,0.7661)(3,0.7664)(4,0.7614)};
\addplot[sensred,only marks,mark=star,mark size=3.6pt] coordinates {(3,0.7664)};
\end{axis}
\begin{axis}[sensright,at={(group c3r1.south west)},anchor=south west,xmin=-0.05,xmax=1.25,ylabel={Permute F1 $\uparrow$}]
\addplot[sensred,mark=square*] coordinates {(0,0.7174)(0.2,0.7634)(0.4,0.7705)(0.6,0.7666)(0.8,0.7679)(1,0.7664)(1.2,0.7661)};
\addplot[sensred,only marks,mark=star,mark size=3.6pt] coordinates {(1,0.7664)};
\end{axis}
\node at ($(group c2r1.north)+(0,0.9cm)$) {\pgfplotslegendfromname{sensitivitylegend}};
\end{tikzpicture}
\caption{Parameter sensitivity on AutoSplice under Protocol~II. F1 and Permute F1 use the left (blue) and right (red) axes, respectively. Stars mark the main settings ($\gamma=1$, $T=3$, $\lambda_{\mathrm{deep}}=1$). Axis ranges vary across panels. Panel (c) includes the zero-weight control.}
\label{fig:parameter_sensitivity}
\end{figure*}

\subsection{Sensitivity to the HEA Scaling Factor}
\label{sec:gamma_sensitivity}

We vary $\gamma\in\{0.5,1,2,4\}$ during both training and inference.
Figure~\ref{fig:parameter_sensitivity}(a) shows similar performance at 0.5 and 1, while larger values reduce both metrics.
For a fixed positive support difference, increasing $\gamma$ gives the candidate more weight, which also amplifies the effect of an incorrect revision.
We use $\gamma=1$ for its highest F1 and near-best Permute F1.

\subsection{Analysis of the Number of Update Steps}
\label{sec:update_step_sensitivity}

Figure~\ref{fig:parameter_sensitivity}(b) shows F1 increasing up to three steps and slightly declining at four.
Permute F1 peaks at $T=1$ but remains close at $T=3$.
The benefit of additional updates saturates around $T=3$, where performance has largely stabilized within the tested range.
Accordingly, we use three steps in the main experiments.

\begin{table*}[!t]
\centering
\caption{Robustness on AutoSplice and SAGI-D-9K under Protocol~II.
Avg.\ F1 is the mean over all seven conditions, including the unperturbed input.
The higher average for each dataset and perturbation is bold.}
\label{tab:robustness}
\resizebox{\textwidth}{!}{%
\begin{tabular}{lllcccccccc}
\toprule
Perturbation & Dataset & Model & \multicolumn{7}{c}{Perturbation strength} & Avg.\ F1 \\
\cmidrule(lr){4-10}
& & & None & 3 / 100 & 7 / 90 & 11 / 80 & 15 / 70 & 19 / 60 & 23 / 50 & \\
\midrule
\multirow{4}{*}{GaussNoise} & \multirow{2}{*}{AutoSplice} & Mesorch & 0.5319 & 0.3699 & 0.2933 & 0.2488 & 0.2169 & 0.1952 & 0.1775 & 0.2905 \\
 &  & Ours & 0.6030 & 0.4576 & 0.3783 & 0.3283 & 0.2956 & 0.2703 & 0.2508 & \textbf{0.3691} \\
\cmidrule(lr){2-11}
 & \multirow{2}{*}{SAGI-D-9K} & Mesorch & 0.7398 & 0.7310 & 0.7246 & 0.7172 & 0.7046 & 0.7014 & 0.6872 & 0.7151 \\
 &  & Ours & 0.7418 & 0.7357 & 0.7277 & 0.7216 & 0.7161 & 0.7100 & 0.7031 & \textbf{0.7223} \\
\midrule
\multirow{4}{*}{GaussianBlur} & \multirow{2}{*}{AutoSplice} & Mesorch & 0.5319 & 0.4193 & 0.3599 & 0.3480 & 0.3278 & 0.2958 & 0.2506 & 0.3619 \\
 &  & Ours & 0.6030 & 0.5029 & 0.4580 & 0.4503 & 0.4295 & 0.3912 & 0.3437 & \textbf{0.4541} \\
\cmidrule(lr){2-11}
 & \multirow{2}{*}{SAGI-D-9K} & Mesorch & 0.7398 & 0.7292 & 0.7024 & 0.6856 & 0.6588 & 0.6358 & 0.6095 & 0.6802 \\
 &  & Ours & 0.7418 & 0.7331 & 0.7068 & 0.6853 & 0.6631 & 0.6407 & 0.6169 & \textbf{0.6839} \\
\midrule
\multirow{4}{*}{JPEGCompression} & \multirow{2}{*}{AutoSplice} & Mesorch & 0.5319 & 0.4747 & 0.2912 & 0.2476 & 0.2978 & 0.3417 & 0.2088 & 0.3420 \\
 &  & Ours & 0.6030 & 0.5493 & 0.3795 & 0.3354 & 0.3842 & 0.4224 & 0.3041 & \textbf{0.4254} \\
\cmidrule(lr){2-11}
 & \multirow{2}{*}{SAGI-D-9K} & Mesorch & 0.7398 & 0.7342 & 0.7167 & 0.7023 & 0.6825 & 0.6718 & 0.6537 & 0.7001 \\
 &  & Ours & 0.7418 & 0.7389 & 0.7216 & 0.7088 & 0.6919 & 0.6814 & 0.6666 & \textbf{0.7073} \\
\bottomrule
\end{tabular}%
}
\par\medskip
\begin{minipage}{\textwidth}
\footnotesize
\textit{Note:} For GaussNoise and GaussianBlur, 3, 7, 11, 15, 19, and 23 denote noise standard deviations and blur kernel sizes, respectively.
For JPEGCompression, the corresponding values 100, 90, 80, 70, 60, and 50 denote JPEG quality factors.
\end{minipage}
\end{table*}

\subsection{Analysis of the Deep Supervision Weight}
\label{sec:deep_weight_sensitivity}

In Figure~\ref{fig:parameter_sensitivity}(c), adding candidate deep supervision substantially improves both metrics over the zero-weight control.
Both peak at $\lambda_{\mathrm{deep}}=0.4$ and remain comparatively stable over the tested range from 0.4 to 1.2.
Performance is therefore less sensitive to the precise weight within this range.
We retain $\lambda_{\mathrm{deep}}=1$ as the default used in the component ablations, both metrics are within half a percentage point of their respective peaks.

\section{Detailed Robustness Results}
\label{sec:supp_robustness}

Table~\ref{tab:robustness} reports the complete perturbation sweep used for the summary in the main paper.
Both Mesorch and our method are trained under Protocol~II and evaluated with the same perturbation settings.
For Gaussian noise and blur, strengths increase from 3 to 23; JPEG quality decreases from 100 to 50.
The unperturbed input is included in every reported average.

Both methods lose F1 as Gaussian noise or blur increases, but our method remains ahead across almost the entire sweep.
On AutoSplice, its advantage extends to every degraded condition; on SAGI-D-9K, the differences are smaller, with one blur setting favoring Mesorch.
The improvement therefore extends beyond clean inputs across the tested severity range.

JPEG performance on AutoSplice is nonmonotonic for both models, so compression quality alone does not predict localization accuracy.
Our method nevertheless maintains higher F1 at every JPEG setting on both datasets.
Overall, the full sweeps show that the benefit of HEA-guided refinement persists under common postprocessing, with degradation still visible in both methods.
These absolute F1 comparisons reflect clean-input accuracy as well as sensitivity to perturbations.

\begin{table*}[!t]
\centering
\small
\setlength{\tabcolsep}{3pt}
\renewcommand{\arraystretch}{1.12}
\caption{Fixed-candidate HEA diagnostics on the native $128\times128$ grid.
(a) Good-candidate selection and realized probability-error changes.
(b) Net mask corrections, corrective-flip shares, and native-grid macro F1.}
\label{tab:hea_mechanism_diagnostic}
\textbf{(a) Candidate selection and realized probability-error changes.}\par\smallskip
\begin{tabular*}{\textwidth}{@{\extracolsep{\fill}}cccccc@{}}
\toprule
\multirow{2}{*}{Step} & \multicolumn{3}{c}{Good proposals (\%)}
& \multirow{2}{*}{\shortstack{HEA--Matched\\(pp; 95\% CI)}}
& \multirow{2}{*}{\shortstack{Realized accepted updates (\%)\\Improved / Worsened / Negligible}} \\
\cmidrule(lr){2-4}
& All & Matched & HEA & & \\
\midrule
1 & 34.66 & 40.17 & 41.42 & +1.26 [1.16, 1.35] & 28.84 / 32.76 / 38.40 \\
2 & 34.62 & 40.88 & 41.09 & +0.21 [0.18, 0.24] & 27.75 / 31.63 / 40.62 \\
3 & 34.63 & 40.95 & 41.07 & +0.12 [0.09, 0.14] & 27.46 / 31.22 / 41.32 \\
\bottomrule
\end{tabular*}
\par\medskip
\textbf{(b) Realized mask corrections and one-step diagnostic F1.}\par\smallskip
\begin{tabular*}{\textwidth}{@{\extracolsep{\fill}}crrcccc@{}}
\toprule
\multirow{2}{*}{Step} & \multicolumn{2}{c}{Net corrected pixels ($10^3$)}
& \multirow{2}{*}{\shortstack{Corrective flips\\(\%)}}
& \multicolumn{3}{c}{Native-grid macro F1} \\
\cmidrule(lr){2-3}\cmidrule(lr){5-7}
& HEA & Matched & & Current $P_{t-1}$ & Direct $Q_t$ & Updated $P_t$ \\
\midrule
1 & 3034.19 & 2274.89 & 80.82 & 0.5315 & 0.6082 & 0.6674 \\
2 & 481.35 & 548.53 & 69.83 & 0.6674 & 0.6563 & 0.6865 \\
3 & 177.55 & 240.98 & 63.12 & 0.6865 & 0.6608 & 0.6937 \\
\bottomrule
\end{tabular*}
\par\smallskip
\begin{minipage}{\textwidth}
\footnotesize
\textit{Notes:} Eligible: $|d_{t,i}|>10^{-3}$; Good: $d_{t,i}>10^{-3}$.
All pools eligible proposals; HEA/Matched use accepted eligible proposals.
Realized outcomes use all $A_{t,i}>0$ pixels: error reductions/increases above $10^{-3}$ are Improved/Worsened, otherwise Negligible.
Net corrections equal corrected minus corrupted labels on eligible pixels; corrective-flip shares use all realized binary flips.
Direct $Q_t$ is a one-step diagnostic without feedback.
Matched expectations redistribute strengths (including zeros) within image, iteration, direction, and current-probability/change-magnitude bins with boundaries $\{0,0.2,0.4,0.6,0.8,1\}$/$\{0,0.01,0.05,0.15,0.35,1\}$.
Selection-lift CIs use 1,500 paired bootstrap resamples of 1,710 filename-prefix groups.
\end{minipage}
\end{table*}

\section{Mechanism Diagnostics of HEA}
\label{sec:supp_hea_mechanism}

\begin{figure*}[!t]
\centering
\includegraphics[width=\textwidth]{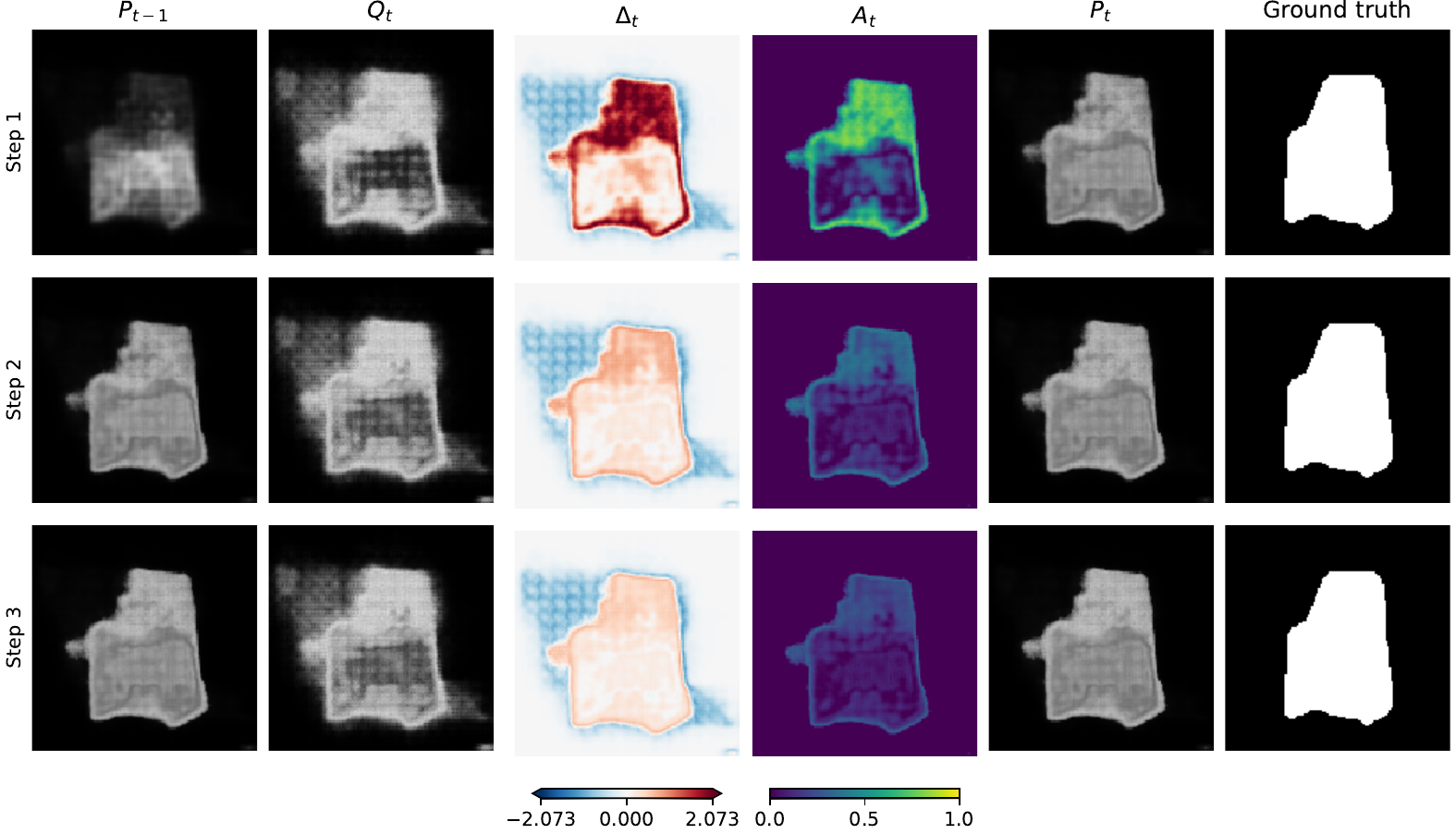}
\caption{Selective adoption of candidate revisions on AutoSplice under Protocol~II.
Columns show the current hypothesis $P_{t-1}$, candidate $Q_t$,
relative support $\Delta_t$, applied update strength $A_t$,
updated hypothesis $P_t$, and GT; rows show three iterations.
This selected successful example illustrates how HEA incorporates
foreground recovery while limiting some candidate-induced changes
to existing responses.
All maps use the native $128\times128$ grid, with nearest-neighbor
GT alignment and shared color scales across iterations.}
\label{fig:hea_62680_trajectory}
\end{figure*}

We evaluate the full model under Protocol~II on all 3,621
AutoSplice images, with $T=3$, $\gamma=1$, and $\alpha=1$.
Diagnostics are computed on the native $128\times128$ update grid,
with nearest-neighbor GT alignment and a binary threshold of 0.5.
At each iteration, we fix the same current--candidate pair
$(P_{t-1},Q_t)$ to examine candidate selection and the resulting update.
The Matched reference preserves the within-group acceptance count
and update-strength distribution.

Table~\ref{tab:hea_mechanism_diagnostic}(a) shows that HEA preferentially selects beneficial proposals.
At the first iteration, the Good-proposal share increases from
34.66\% before selection to 41.42\% after HEA acceptance,
exceeding the matched-budget expectation of 40.17\%.
The additional gain over Matched is largest at this initial step.
Accepted updates nevertheless show mixed probability-level changes:
worsening pixels outnumber improving pixels, but these counts
do not weight change magnitudes or require binary-mask flips.
They therefore measure a different aspect of refinement from
the mask-level corrections in Table~\ref{tab:hea_mechanism_diagnostic}(b).

At the binary-mask level, 80.82\%, 69.83\%, and 63.12\%
of the realized label flips correct an existing error in the
three iterations, respectively.
This corrective-flip share counts FN-to-TP and FP-to-TN
transitions relative to all realized binary-label changes.
Moreover, soft updating achieves higher native-grid F1 than
directly adopting the same candidate at every step.
This is a fixed-pair, one-step comparison without recursive feedback.
These results support selective, partial adoption as an effective
way to retain useful candidate revisions while avoiding some errors
introduced by complete replacement.

The refinement trajectory also becomes progressively more stable.
The number of realized label changes decreases from 4.922 million
to 1.214 million and then 0.677 million, while the incremental
F1 gains decrease from 13.59 to 1.91 and 0.72 percentage points.
The negligible probability-change share also rises from
38.40\% to 41.32\%.
Correspondingly, the updated F1 increases from 0.6674 to
0.6865 and 0.6937.
Together, these observations suggest that the initial iteration
performs the major corrections, followed by smaller adjustments
as the hypothesis stabilizes.
Although individual revisions remain imperfect, the evaluated
trajectory shows sustained improvement with diminishing changes
across iterations.
These are native-grid diagnostics, distinct from the
full-resolution evaluation in the main paper.

To complement the aggregate diagnostics, Figure~\ref{fig:hea_62680_trajectory} traces
candidate revisions, relative support, and applied update strengths
to show how HEA selectively changes the localization hypothesis.

\newpage
\paragraph{Qualitative example.}
Figure~\ref{fig:hea_62680_trajectory} illustrates selection within a single candidate revision.
The first candidate recovers the missing upper region but also
weakens some interior responses.
HEA applies stronger updates to the recovered upper region
and parts of the outline, while assigning lower strengths
to the interior changes.
The resulting $P_1$ achieves a native-grid F1 of 0.8478,
compared with 0.7287 from directly adopting $Q_1$.
This comparison illustrates the benefit of incorporating useful
candidate changes without fully replacing the current hypothesis.
Later iterations make smaller adjustments as the support differences
and update strengths decrease, yielding a final F1 of 0.8794,
up from 0.4854 at $P_0$.
The final localization is more complete, with some boundary
mismatches remaining.

Together, the aggregate diagnostics and the traced example support
HEA's preference for beneficial revisions and the value of selective
soft adoption, with the strongest additional selection benefit
at the initial iteration.
Subsequent updates progressively stabilize the hypothesis through
smaller revisions, complementing the end-to-end ablation results.

\end{document}